\documentclass[letterpaper,10pt,conference]{ieeeconf}
\IEEEoverridecommandlockouts
\usepackage{graphicx,amsmath,amssymb,booktabs,multirow,cite,url,array,xcolor,colortbl,tabularx,type1cm}
\usepackage{flushend,flafter}
\usepackage[protrusion=false]{microtype}
\usepackage{caption}
\usepackage{hyperref}
\hypersetup{hidelinks}

\newcommand{\sensor}{GelSlim~5.0}
\newcommand{\method}{VisTacFusion}
\newcolumntype{L}{>{\raggedright\arraybackslash}X}
\newcommand{\tableformat}{\small\setlength{\tabcolsep}{3.0pt}\renewcommand{\arraystretch}{1.02}}
\newcommand{\metrichead}[2]{\shortstack[c]{\strut#1\\\strut#2\,$\downarrow$}}
\title{\LARGE\bf GelSlim 5.0: Unifying Vision and Touch with a Single Fingertip Camera}
\author{Xiangdong Wang$^{*}$, I-Hsuan Huang$^{*}$, Xiaohao Xu$^{*}$, Xili Yi, Nima Fazeli, and Xiaonan Huang
\thanks{$^{*}$Equal contribution. All authors are with the University of Michigan, Ann Arbor, MI, USA. Correspondence: \texttt{\{sheldonw, ihsuan, xiaohaox\}@umich.edu}.}}
\hypersetup{
    pdftitle={GelSlim 5.0: Unifying Vision and Touch with a Single Fingertip Camera},
    pdfauthor={Xiangdong Wang, I-Hsuan Huang, Xiaohao Xu, Xili Yi, Nima Fazeli, Xiaonan Huang}
}
\begin{document}
\maketitle
\thispagestyle{empty}\pagestyle{empty}\raggedbottom
\begin{abstract}
Vision and touch provide complementary cues for object pose and local contact geometry, but external cameras are often occluded at the contact site. We present \sensor{}, a fingertip sensor that acquires visual and tactile observations with one fixed camera and a continuous UV-responsive contact layer. By switching illumination, the camera sequentially acquires through-membrane visual images and contact-sensitive tactile images from a shared viewpoint, avoiding inter-camera extrinsic calibration. To combine these observations, we introduce \method{}, a multimodal transformer with separate prediction paths for planar object pose and contact geometry. Asymmetric bottleneck fusion incorporates visual context while preserving tactile spatial features for depth and surface-normal estimation. Modality dropout enables inference with paired inputs or either modality alone. On real validation data from 12 known objects, paired inputs reduce depth MSE, angular error, and translation $L_1$ by 61.2\%, 72.3\%, and 61.3\%, respectively, relative to tactile-only inference with the same model. Relative to visual-only inference, the corresponding reductions are 93.7\%, 6.1\%, and 8.6\%. \href{https://gelslim5.github.io/geslim5_website/}
{\textcolor{blue}{Project website}}
\end{abstract}

\section{INTRODUCTION}
\begin{figure}[t]
\centering
\includegraphics[width=\columnwidth]{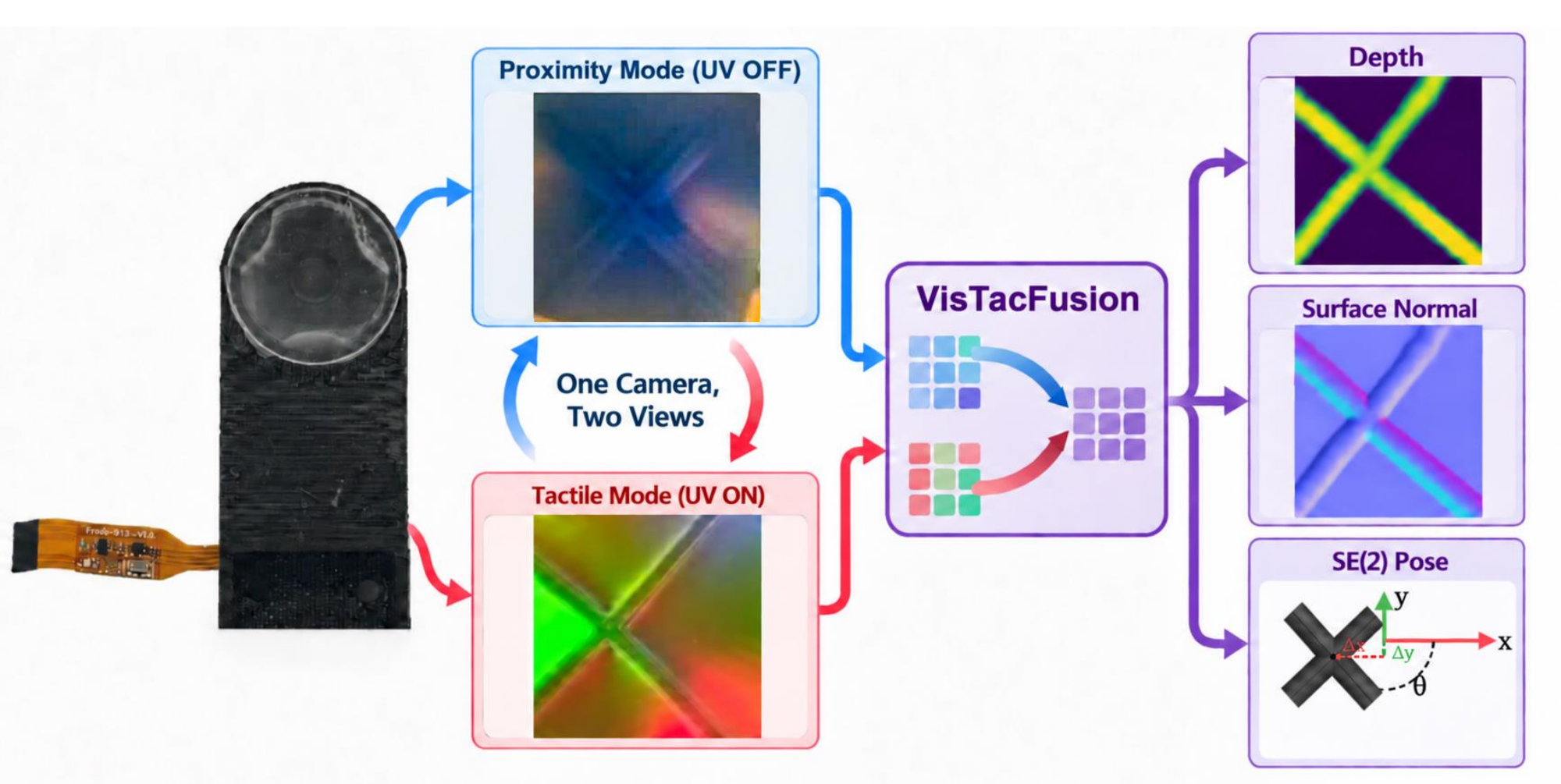}
\caption{\textbf{\sensor{} enables visual and tactile feedback with a single fingertip camera.} By switching illumination, the sensor operates in proximity mode with UV off, allowing the camera to see through the membrane, and in contact-sensitive tactile mode with UV on. Both modes share the same viewpoint and require no inter-camera extrinsic calibration. \method{} combines their complementary signals to estimate planar object rotation and translation, as well as contact depth and surface normals. The observations are acquired sequentially under approximately stationary contact.}
\label{fig:teaser}
\end{figure}

Contact-rich manipulation requires estimating both object pose and the local geometry of contact. Vision supplies object context, while optical tactile sensors such as GelSight and GelSlim capture detailed contact structure~\cite{yuan2017gelsight,donlon2018gelslim}. The modalities can resolve different ambiguities: visual context can distinguish object poses that produce similar contact patches, while tactile measurements reveal local surface geometry that visual appearance alone may not resolve.\unskip

With external cameras, however, the hand and object can occlude the contact site, and viewpoint differences complicate spatial association with tactile measurements. A shared fingertip camera places visual sensing at the contact interface and provides a common frame for relating object appearance to contact geometry. Prior visuotactile sensors obtain scene access through semitransparent skins~\cite{hogan2021sts}, controllable films~\cite{luu2023protac}, UV-visible markers~\cite{wang2022spectac}, or spatial multiplexing~\cite{hu2026muxgel}.
Building on these designs, we couple a continuous contact interface with an estimator for object pose and spatially resolved contact geometry.\unskip

We introduce \sensor{}, a fingertip sensor that co-locates vision and touch within the GelSlim~4.0 body design~\cite{sipos2024gelslim4}. A continuous UV-responsive contact layer allows one fixed camera to record through-membrane visual images with UV off and contact-sensitive tactile images with UV on (Fig.~\ref{fig:teaser}).
The observations are acquired sequentially under approximately stationary contact and share a viewpoint and image coordinates, supporting spatial association without inter-camera extrinsic calibration.
Table~\ref{tab:sensor_design_comparison} positions the sensor among prior interfaces.

\begin{table*}[t]
\centering
\begingroup
\fontsize{7.0}{7.9}\selectfont
\setlength{\tabcolsep}{1.5pt}
\renewcommand{\arraystretch}{1.03}
\setlength{\aboverulesep}{1.4pt}
\setlength{\belowrulesep}{1.4pt}
\setlength{\extrarowheight}{0pt}
\begin{tabular*}{\textwidth}{@{\extracolsep{\fill}}
 >{\raggedright\arraybackslash}p{21mm}
 >{\centering\arraybackslash}p{10mm}
 >{\raggedright\arraybackslash}p{22mm}
 >{\raggedright\arraybackslash}p{30mm}
 >{\raggedright\arraybackslash}p{17mm}
 >{\raggedright\arraybackslash}p{10mm}
 >{\raggedright\arraybackslash}p{18mm}
 >{\raggedright\arraybackslash}p{23mm}
 >{\raggedright\arraybackslash}p{16mm}@{}}
\toprule
\multicolumn{4}{c}{\textbf{Sensor construction and physical size}} &
\multicolumn{2}{c}{\textbf{Mode selection}} &
\multicolumn{3}{c}{\textbf{Observation and reconstruction}}\\
\cmidrule(lr){1-4}\cmidrule(lr){5-6}\cmidrule(lr){7-9}
\textbf{Sensor} & \textbf{Cameras} & \textbf{Contact interface} &
\textbf{Size [mm]}\newline H: head; I: interface &
\textbf{Selected element} & \textbf{Control} & \textbf{Tactile cue} &
\textbf{Contact representation} & \textbf{Acquisition}\\
\midrule
\multicolumn{9}{@{}l}{\itshape Persistent transparency or parallel spectral channels}\\[1pt]
ViTacTip~\cite{fan2024vitactip,zhang2025multimodality} & 1 shared & Clear skin; marker-tipped pins &
H: NR\newline I: $2500\,\mathrm{mm}^{2\,\mathrm{S}}$ &
None & None & Pin / marker motion & Marker deformation & Same frame\\
M\textsuperscript{3}Tac~\cite{li2024m3tac} & 3 bands$^{a}$ & Spectrally selective membrane &
H: $\varnothing65\!\times\!80^{\mathrm{R}}$\newline I: film $t=0.2^{\mathrm{R}}$; area NR &
Spectral channels & Filters + LEDs & Photometric shading & Surface reconstruction & Parallel bands\\
\midrule
\multicolumn{9}{@{}l}{\itshape Fixed interface with illumination-dependent observations}\\[1pt]
STS~\cite{hogan2021sts} & 1 shared & Semitransparent reflective skin &
H: NR\newline I: $150\!\times\!150^{\mathrm{R}}$ pad ($t=5$) &
Lighting ratio & LEDs & Photometric shading & Contact images & Alternating / mixed\\
StereoTac~\cite{roberge2023stereotac} & 2 stereo & Semitransparent reflective skin &
H: NR\newline I: square; side NR &
Illumination & LEDs & Photometric shading & Surface depth & Alternating\\
VisTac~\cite{athar2023vistac} & 2 paths & Semitransparent reflective skin &
H: $70\!\times\!38\!\times\!33^{\mathrm{R}}$\newline I: $45\!\times\!34^{\mathrm{R}}$ pad;\newline \hspace*{1.5mm}$32\!\times\!32^{\mathrm{R}}$ active &
Illumination & LEDs & Photometric shading & Surface depth & Sequential\\
SpecTac~\cite{wang2022spectac} & 1 shared & Clear gel; fluorescent markers &
H: NR\newline I: $\varnothing$ NR; $t\approx10+0.5^{\mathrm{R}}$ &
Marker visibility & UV LEDs & Marker motion & Marker deformation & UV-selected\\
TransTac~\cite{yang2026transtac} & 2 stereo & Clear gel; UV markers &
H: NR\newline I: $t=5^{\mathrm{R}}$; lateral NR &
Marker visibility & UV LEDs & Stereo marker motion & 3D marker geometry & UV-selected\\
\midrule
\multicolumn{9}{@{}l}{\itshape Auxiliary transparency control or mechanical interface selection}\\[1pt]
ProTac~\cite{luu2023protac} & 2 views & Clear skin; PDLC film; markers &
H: NR; I: NR &
Auxiliary PDLC & Voltage & Marker motion & Skin deformation & Alternating\\
Look-to-Touch~\cite{dong2025looktotouch} & 1 shared & Clear / opaque belt regions &
H: NR; I: NR &
Exposed belt region & Servo & Photometric shading & Surface depth & Sequential\\
\midrule
\multicolumn{9}{@{}l}{\itshape No switching -- GelSlim family}\\[1pt]
GelSlim (1.0)~\cite{donlon2018gelslim} & 1 tactile & Opaque reflective skin &
H: NR\newline I: $50\!\times\!50^{\mathrm{R}}$ pad &
None & None & Photometric shading & Contact images & Tactile only\\
GelSlim 2.0~\cite{ma2019gelslim2,taylor2022gelslim3} & 1 tactile & Opaque skin; markers &
H: $50\!\times\!172\!\times\!25^{\mathrm{S}}$\newline I: $1200\,\mathrm{mm}^{2\,\mathrm{S}}$ &
None & None & Shading + marker motion & Marker deformation$^{b}$ & Tactile only\\
GelSlim 3.0~\cite{taylor2022gelslim3} & 1 tactile & Opaque skin; markers &
H: $37\!\times\!80\!\times\!20^{\mathrm{R}}$\newline I: $675\,\mathrm{mm}^{2\,\mathrm{R}}$ &
None & None & Shading + marker motion & Surface depth & Tactile only\\
GelSlim 4.0~\cite{sipos2024gelslim4} & 1 tactile & Opaque skin; optional markers &
Body: $83.6\!\times\!20.1\!\times\!35.1^{\mathrm{C}}$\newline I: NR &
None & None & Photometric shading & Surface depth & Tactile only\\
\midrule
\textbf{GelSlim 5.0 (ours)} & \textbf{1 shared} & \textbf{Continuous UV-responsive film}$^{c}$ &
$83.6\!\times\!22.1\!\times\!35.1^{\mathrm{A,C}}$\newline \textbf{UV layer}; I: NR &
\textbf{Contact-film response} & \textbf{UV LEDs} & \textbf{Contact intensity / shading} & \textbf{Learned depth + normals} & \textbf{Sequential}\\
\bottomrule
\end{tabular*}
\par\vspace{1pt}
\begin{minipage}{\textwidth}
\fontsize{6.4}{7.3}\selectfont
\textit{Size guide.} H denotes the reported sensing-head envelope; I denotes the contact/active interface; NR means not reported; $t$ is thickness.
Dimensions retain the reported axis order; areas are explicitly in mm$^2$.
$^{\mathrm{R}}$Primary-paper value; $^{\mathrm{S}}$later scholarly report; $^{\mathrm{C}}$official body-CAD bounding box; $^{\mathrm{A}}$author-provided design constraint.
GelSlim~2.0 dimensions refer to the standard prototype summarized in GelSlim~3.0.
GelSlim~4.0 reports the gripper-body CAD. GelSlim~5.0 adds the author-specified UV layer along the 20.1\,mm axis, giving a nominal $83.6\!\times\!22.1\!\times\!35.1$\,mm body-plus-layer envelope. Neither value describes a complete assembly or contact area; separate back covers, cables, and robot adapters are excluded.
SpecTac lists gel and protective-layer \mbox{thickness separately, not their summed thickness. A platform pad, camera baseline, or belt stock is not a head envelope.}\unskip\par
\textit{Reading guide.} Geometry refers to the contact interface, not external-scene depth. Output labels describe cited demonstrations rather than limits on other possible outputs; photometric methods may estimate normals or gradients internally.
``UV-selected'' does not assert same-exposure separation.
$^{a}$Visible, near-infrared, and mid-infrared imagers, not three RGB cameras.
$^{b}$GelSlim~2.0 also estimates force distributions using inverse FEM.
$^{c}$UV-responsive denotes the observed imaging-state change; transmission, reflection, and UV emission have not been \mbox{separately quantified. Our reconstruction is evaluated on known-object, within-session validation samples.}\unskip
\end{minipage}
\endgroup
\caption{\textbf{Representative sensor designs and physical sizes for scene and contact perception.} Rows summarize cited implementations, not a common accuracy benchmark.\unskip}
\label{tab:sensor_design_comparison}
\end{table*}

To exploit these complementary observations, \method{} uses separate prediction paths for planar object pose and contact geometry. Tactile-initialized queries provide a pose estimate, while intermediate tactile features preserve spatial detail for contact-depth and surface-normal estimation. A shared visual bottleneck conditions both paths, allowing object context to inform pose and geometry predictions. Modality dropout trains the model to operate with either input alone or the pair.

Our contributions are \textbf{(1)} a fingertip sensor that co-locates vision and touch using a continuous UV-responsive contact interface; \textbf{(2)} an estimator that combines visual context with separate pose queries and spatial geometry features; and \textbf{(3)} an evaluation on 12 known objects covering modality complementarity, encoder and fusion choices, and training configuration. Within-session validation shows lower contact-depth and pose errors for paired observations than for either input alone in the same estimator.\unskip



\section{RELATED WORK}
\label{sec:related_work}
\begin{figure}[t]
\centering
\includegraphics[width=\columnwidth]{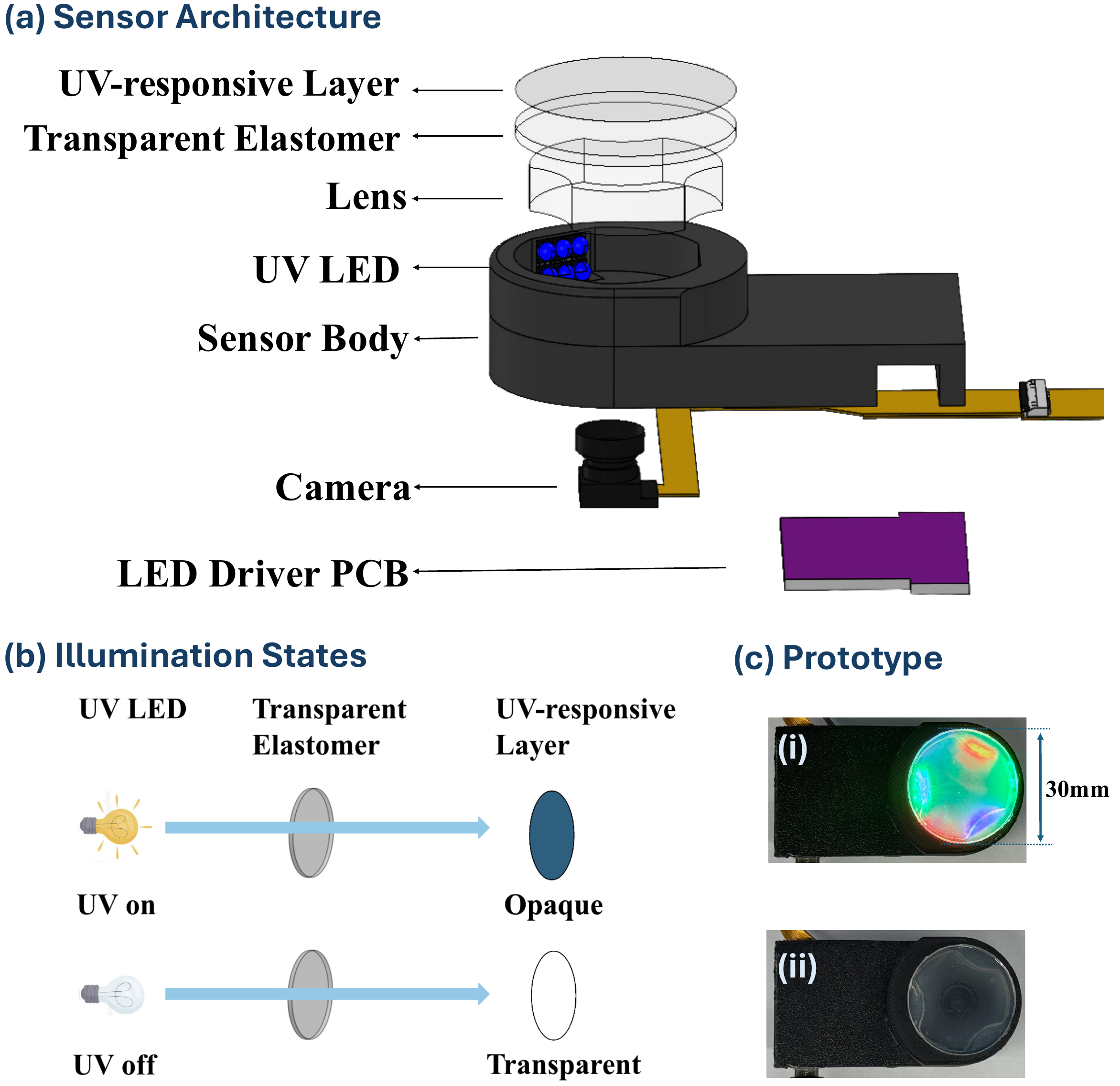}
\caption{\textbf{\sensor{} hardware and observation states.} (a) Exploded assembly with the UV-responsive film at the contact interface. (b) Illumination-controlled visual and tactile acquisition. (c) Prototype in (i) tactile mode and (ii) proximity mode; scale marker: 30\,mm. \unskip}
\label{fig:sensor}
\end{figure}
\noindent\textbf{From tactile fingers to switchable visuotactile sensing.} GelSight, DIGIT, and the GelSlim family image a deformable contact interface~\cite{yuan2017gelsight,lambeta2020digit,donlon2018gelslim,ma2019gelslim2,taylor2022gelslim3,sipos2024gelslim4}. GelSlim~4.0 emphasizes reproducible hardware and perception~\cite{sipos2024gelslim4}; \sensor{} retains its body design and changes the contact film and illumination to add a visual observation state. Scene access has also been obtained with semitransparent interfaces in STS, VisTac, and StereoTac~\cite{hogan2021sts,athar2023vistac,roberge2023stereotac}; UV-visible markers in SpecTac and TransTac~\cite{wang2022spectac,yang2026transtac}; PDLC control in ProTac~\cite{luu2023protac}; and mechanical switching in Look-to-Touch~\cite{dong2025looktotouch}. ViTacTip and M$^3$Tac use transparent or spectrally selective constructions~\cite{fan2024vitactip,li2024m3tac}. MuxGel spatially interleaves transparent and tactile regions for single-camera reconstruction~\cite{hu2026muxgel}, while FingerEye uses binocular visuotactile sensing for manipulation~\cite{xu2026fingereye}. Our implementation combines a continuous UV-responsive outer contact film, sequential same-camera acquisition, and learned pose/contact estimation within the GelSlim body. The estimator uses contact-sensitive image appearance rather than marker displacement, without spatially partitioning the film into clear and coated windows. Table~\ref{tab:sensor_design_comparison} compares construction, acquisition, and contact representation; head dimensions and contact-interface measurements are listed \mbox{separately within the size column.}\unskip

\vspace{2pt}
\noindent\textbf{Learning from visual and tactile representations}
Visual pretraining methods such as MAE, DINO, Depth Anything, CLIP, and SigLIP provide distinct feature representations~\cite{he2022mae,oquab2023dinov2,simeoni2025dinov3,yang2024dav2,radford2021clip,zhai2023siglip}. Tactile and cross-modal representations include Sparsh, T3, SITR, TVL, and Sparsh-X~\cite{higuera2024sparsh,zhao2024transferable,gupta2025sitr,fu2024tvl,higuera2025sparshx}, while MViTac learns visual--tactile alignment by contrastive pretraining~\cite{dave2024mvitac}. We use these representations for object-pose and contact-geometry estimation, rather than introducing a new backbone-pretraining objective. Building on attention bottlenecks~\cite{nagrani2021mbt} and DPT-style dense decoding~\cite{ranftl2021dpt}, \method{} uses compact visual context to condition a pose path and a separate contact-geometry decoder. Simulators such as TACTO motivate supplementing real tactile data with synthetic observations~\cite{wang2022tacto}; our evaluation uses Blender-based co-training, not simulation-only transfer.\unskip

\section{GELSLIM 5.0: COMMON-VIEW SENSING}
\label{sec:sensor}

\begin{figure*}[t]
\centering
\includegraphics[width=\textwidth]{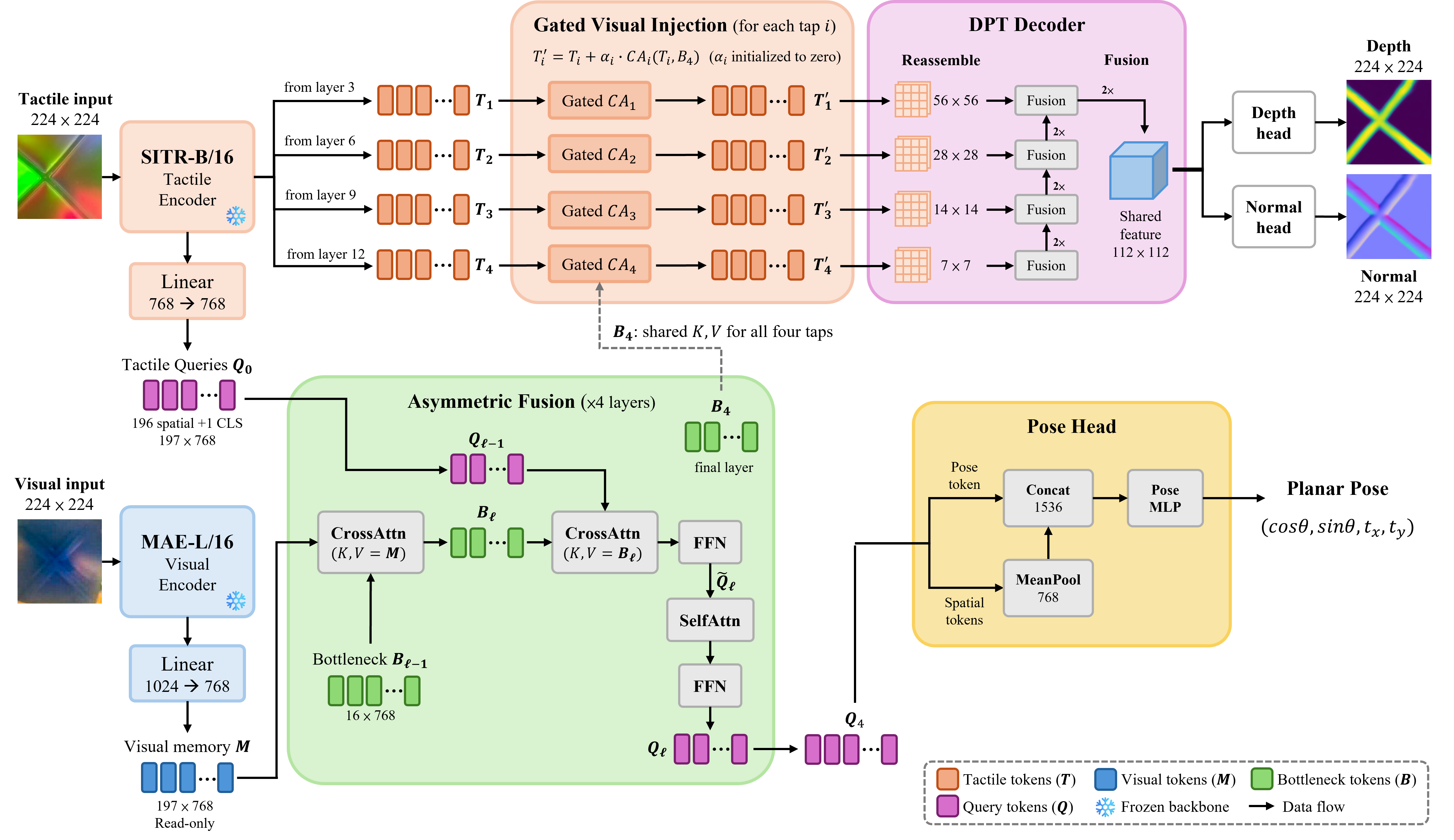}
\caption{\textbf{\method{}: task-specific estimation of object pose and contact geometry.} Frozen encoders provide tactile features and visual memory $M$. The pose path uses a 16-token visual bottleneck to condition tactile-initialized queries over four layers; the final query state $Q_4$ predicts planar pose. The geometry path retains tactile maps from blocks 3, 6, 9, and 12, applies gated conditioning from visual-context state $B_4$, and decodes contact depth and surface normals with DPT. Calibration inputs, embeddings, and pre-normalized residual connections are omitted. Geometry \mbox{thumbnails illustrate estimation targets; missing-input routing is defined in Sec.~\ref{sec:three_configs}.}\unskip}
\label{fig:architecture}
\end{figure*}

\sensor{} is a monocular visual--tactile fingertip sensor that captures both observation modalities from a single camera by switching UV illumination (Fig.~\ref{fig:sensor}). The design has three elements: (i)~a compliant contact membrane whose optical state changes under UV, (ii)~an illumination ring that drives the switching, and (iii)~a 3D-printed body housing one camera and a custom driver PCB. Below, we describe each element and the resulting acquisition protocol.

\noindent\textbf{Compliant contact interface.} The prototype retains the GelSlim~4.0 gripper-body CAD~\cite{sipos2024gelslim4} and uses a two-layer Solaris (Smooth-On) silicone membrane. We form a 5\,mm clear base by mixing the silicone 1:1 by weight, vacuum-degassing, and casting it in a mold. The continuous UV-responsive layer is spin-coated from 3\,g Solaris mixed with 0.3\,mL invisible UV-reactive ink (United Nuclear). This layer contacts the object directly, without a separate protective overlayer.\unskip

The 3D-printed PLA body has a nominal circular sensing area of 30\,mm diameter and houses a Raspberry Pi camera. An illumination ring uses UV LEDs in place of the blue channel; red and green illumination remain available. A custom driver PCB and a GPIO-controlled MOSFET switch the UV channel and coordinate image capture. \unskip



\noindent\textbf{Visual and tactile acquisition} With UV off, the camera records a through-membrane visual image $I_v$; we refer to this state as proximity mode. With UV on, it records a contact-sensitive tactile image $I_t$. We use UV-responsive to describe this observed change in imaging state; transmission, reflection, and UV-induced emission have not been separately quantified.\unskip

Each pair is captured sequentially under approximately stationary contact.
Because the camera and lens remain fixed, the images share a viewpoint and image coordinates, supporting spatial association of appearance and contact features without inter-camera extrinsic calibration (Fig.~\ref{fig:real_obj}). Refraction, membrane deformation, and residual motion can still affect pixelwise alignment. Camera intrinsics and object-pose calibration remain relevant to estimation.\unskip



\section{LEARNING OBJECT POSE AND CONTACT GEOMETRY} \label{sec:method}

\method{} combines visual context with tactile signatures through separate pose and geometry prediction paths (Fig.~\ref{fig:architecture}).
The pose path aggregates features into a pose vector, while the geometry path preserves spatial detail for depth and normals. Cross-attention associates the modalities without requiring pixelwise correspondence.

\subsection{Problem Formulation} \label{sec:problem}

For a known object $o$, let $I_v$ and $I_t$ denote the image pair acquired under the quasi-static protocol of Sec.~\ref{sec:sensor}. The input $\mathcal I_m$ is $I_v$, $I_t$, or $(I_v,I_t)$ for modes $m\in\{\mathrm{Vis},\mathrm{Tac},\mathrm{Both}\}$, respectively. We learn an estimator with parameters $\phi$,
\begin{equation}
 f_\phi(\mathcal I_m,o)=\bigl(\widehat{\mathbf y},\widehat D,\widehat{\mathbf N}\bigr),
 \label{eq:problem}
\end{equation}
where $\mathbf y=(\cos\theta,\sin\theta,t_x,t_y)$ encodes sensor-relative planar pose. Depth $D\in\mathbb R^{H\times W}$ and surface normals $\mathbf N\in\mathbb R^{H\times W\times3}$ describe contact geometry on the tactile image grid. Translations are normalized as in Sec.~\ref{sec:setup}; object identity $o$ is supplied to the pose queries.



\subsection{Modality-Specific Features}
\label{sec:encoders}
Frozen SITR-B/16~\cite{gupta2025sitr} and MAE-L/16~\cite{he2022mae} encode touch and vision, respectively. At $224\times224$ resolution, each produces 196 patch tokens and one CLS token, with widths 768 and 1024. SITR uses a fixed set of 18 calibration references; calibration tokens are discarded before downstream prediction.

Separate linear projections map both streams to width 768 and add modality embeddings. Tactile patch tokens receive spatial embeddings and initialize 196 spatial queries; the projected CLS token initializes one pose query. All queries receive an object-identity embedding. Projected visual tokens form fixed memory $M\in\mathbb R^{197\times768}$. All downstream modules are trained jointly.

\subsection{Object Pose Estimation}
\label{sec:trunk}
Tactile-initialized queries $Q_0\in\mathbb R^{197\times768}$ read visual context through 16 learned bottleneck tokens $B_0\in\mathbb R^{16\times768}$. Define the pre-normalized attention residual
\begin{equation}
\mathcal R(X,Y)=X+\operatorname{MHA}(\operatorname{LN}_q(X),\operatorname{LN}_{kv}(Y)),
\end{equation}
where $X$ supplies queries and $Y$ supplies keys and values, and let $\mathcal F(X)=X+\operatorname{FFN}(\operatorname{LN}(X))$. Four layers, indexed by $\ell\in\{1,\ldots,4\}$, update
\begin{align}
B_\ell&=\mathcal R_{B,\ell}(B_{\ell-1},M),\\
\widetilde Q_\ell&=\mathcal F_{1,\ell}\bigl(\mathcal R_{Q,\ell}(Q_{\ell-1},B_\ell)\bigr),\\
Q_\ell&=\mathcal F_{2,\ell}\bigl(\mathcal R_{S,\ell}(\widetilde Q_\ell,\widetilde Q_\ell)\bigr).
\end{align}
Each layer updates the bottleneck from visual memory, then applies query-to-bottleneck cross-attention and query self-attention. States $B_\ell$ and $Q_\ell$ carry across layers; only $B_0$ is a shared learned initialization. Attention uses eight heads. Query FFNs expand the width fourfold with GELU and dropout; the bottleneck update has no FFN.

Writing $Q_4=[s_{4,1},\ldots,s_{4,196},p_4]$, the pose head concatenates the final pose token with the mean of the 196 spatial tokens
\begin{equation}
z_{\rm pose}=[p_4;\operatorname{mean}_{j}s_{4,j}]\in\mathbb R^{1536}.
\end{equation}
A layer-normalized three-layer MLP maps this vector to two rotation components and two translations. The rotation components are normalized to form $(\cos\hat\theta,\sin\hat\theta)$.

\subsection{Contact Geometry Estimation}
\label{sec:decoupled}
To preserve local contact structure, the geometry path uses intermediate tactile features directly. Four feature maps from encoder blocks 3, 6, 9, and 12 bypass the pose-query trunk. After adding position embeddings, each map $T_i\in\mathbb R^{196\times768}$ receives context from the final visual bottleneck $B_4$:
\begin{equation}
T'_i=T_i+\alpha_i\,\operatorname{CA}_i(T_i,B_4),\qquad i=1,\ldots,4.
\label{eq:tap_inject}
\end{equation}
Each map has a separate pre-normalized cross-attention module $\operatorname{CA}_i$, with $T_i$ as queries and $B_4$ as key-values. The learned scalar gates $\alpha_i$ are initialized as zero, so visual conditioning is learned as a residual update to the tactile features. Both pose and geometry losses supervise the shared visual summary $B_4$.
The geometry decoder uses the conditioned spatial features directly, without depending on pose predictions. 

A DPT-style decoder~\cite{ranftl2021dpt} reassembles the four conditioned maps into 256-channel maps at $56^2$, $28^2$, $14^2$, and $7^2$ resolutions. Coarse-to-fine fusion produces a shared $112\times112$ feature map. Separate convolutional heads predict depth and surface normals at $224\times224$; predicted normal vectors are not explicitly normalized.




\subsection{Input Availability and Joint Training}
\label{sec:three_configs}
Modality dropout trains the same model with paired or individual inputs.
The Both mode follows the architecture above. In \emph{Vis}, learned mask queries replace tactile queries, and MAE maps from blocks 3, 6, 9, and 12 (projected from 1024 to 768 channels) supply the geometry features. In \emph{Tac}, both fusion cross-attention updates and the geometry path's visual conditioning are skipped.

All three modes receive pose and geometry supervision.
For inputs that include vision, geometry-path visual conditioning is independently enabled with probability $p_{\rm inject}$ during training and always enabled at inference.



Depth and normal losses are full-image MSEs. Rotation uses $1-\cos(\hat\theta-\theta)$, and translation uses coordinate-averaged $L_1$ error in normalized units. For $k\in\{D,\mathbf N,\mathrm{rot},\mathrm{trans}\}$, the uncertainty-weighted training objective~\cite{kendall2018multi} is
\begin{equation}
\mathcal L=\sum_k\left(e^{-s_k}\mathcal L_k+\tfrac12s_k\right),
\end{equation}
where each learned scalar $s_k$ weights one task.
\section{EXPERIMENTAL EVALUATION}
\label{sec:experiments}
We evaluate modality complementarity, encoder selection, fusion design, and training-data composition under a shared real-data validation protocol.
\begin{figure}[t]
\centering
\includegraphics[width=\columnwidth]{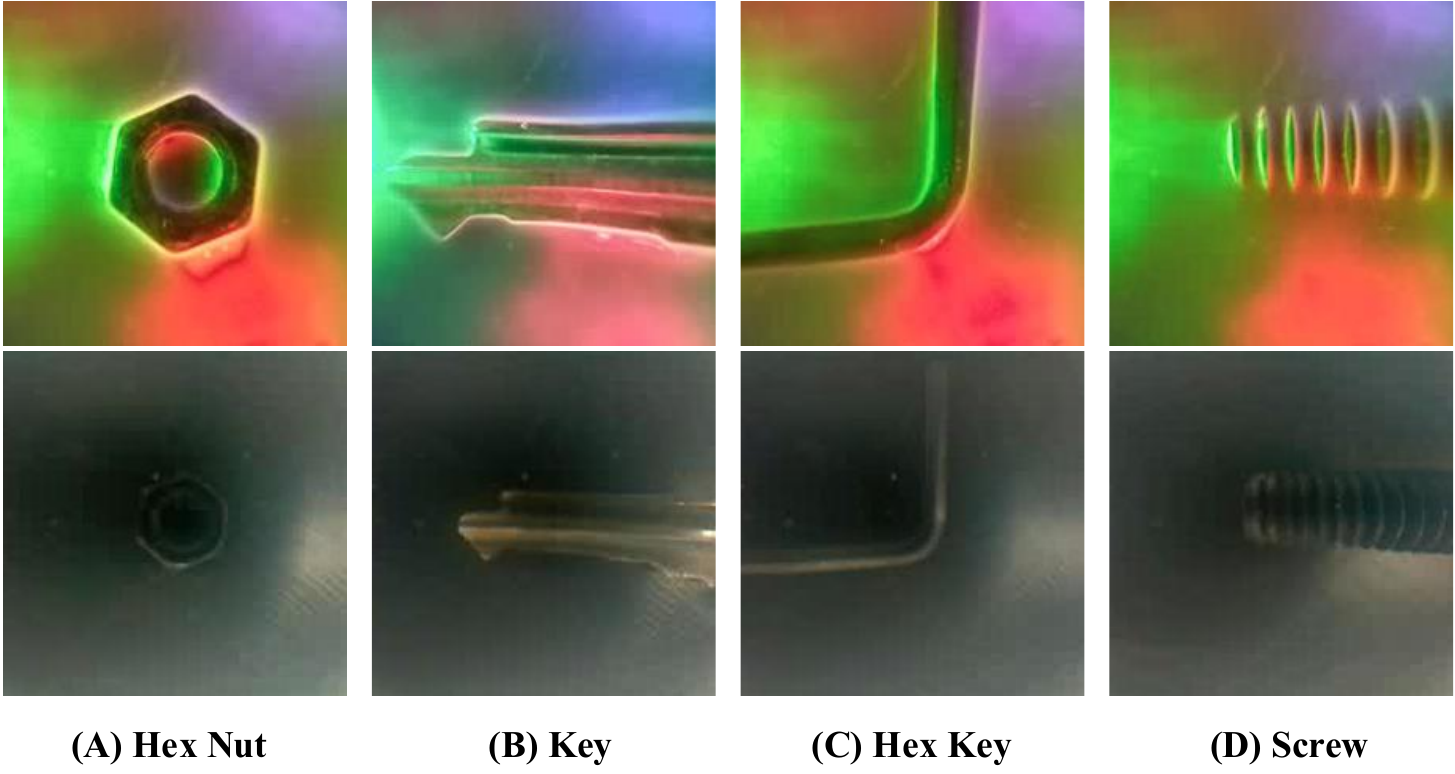}
\caption{\textbf{Common-view visual--tactile observations.} UV-on tactile images (top) and UV-off proximity images (bottom) are acquired sequentially from the same fixed camera under quasi-static contact.}
\label{fig:real_obj}
\end{figure}
\begin{table}[!t]
\centering
\tableformat\setlength{\tabcolsep}{1.5pt}
\begin{tabularx}{\columnwidth}{Lrrrrr}
\toprule
 & \multicolumn{2}{c}{\textbf{Geometry}}
 & \multicolumn{3}{c}{\textbf{Pose}} \\
\cmidrule(lr){2-3} \cmidrule(lr){4-6}
Model & \metrichead{Depth}{MSE} & \metrichead{Normal}{MSE}
 & \metrichead{Rot.}{$L_1$} & \metrichead{Angle}{$({}^\circ)$} & \metrichead{Trans.}{$L_1$} \\
\midrule
\multicolumn{6}{@{}l}{\textit{(a) Vis only}} \\
MAE~\cite{he2022mae}
 & 0.0756 & 0.0403 & 0.0135 & 0.678 & 0.0060 \\
DINOv3~\cite{simeoni2025dinov3}
 & 0.0968 & 0.0358 & 0.0256 & 1.260 & 0.0110 \\
CLIP~\cite{radford2021clip}
 & 0.1501 & 0.0380 & 0.0554 & 2.554 & 0.0294 \\
SigLIP~\cite{zhai2023siglip}
 & 0.1449 & 0.0392 & 0.0525 & 2.526 & 0.0277 \\
T3~\cite{zhao2024transferable}
 & 0.0505 & 0.0250 & 0.0135 & 0.671 & 0.0059 \\
DAv2~\cite{yang2024dav2}
 & 0.0678 & 0.0275 & 0.0141 & 0.701 & 0.0060 \\
\addlinespace[1.5pt]
TVL + MAE$^{\dag}$~\cite{fu2024tvl,he2022mae}
 & 0.0523 & 0.0322 & 0.0156 & 0.772 & 0.0060 \\
Sparsh-X+MAE$^{\dag}$~\cite{higuera2025sparshx,he2022mae}
 & 0.0561 & 0.0290 & 0.0145 & 0.723 & 0.0059 \\
\rowcolor{black!5}
\textbf{\method{}}$^{\dag}$
 & 0.1055 & 0.0388 & 0.0138 & 0.694 & 0.0058 \\
\midrule
\multicolumn{6}{@{}l}{\textit{(b) Tac only}} \\
DAv2~\cite{yang2024dav2}
 & 0.0167 & 0.0128 & 0.0474 & 2.279 & 0.0116 \\
Sparsh-X~\cite{higuera2025sparshx}
 & 0.0183 & 0.0138 & 0.0474 & 2.268 & 0.0119 \\
DINOv3~\cite{simeoni2025dinov3}
 & 0.0174 & 0.0136 & 0.0663 & 3.185 & 0.0162 \\
T3~\cite{zhao2024transferable}
 & 0.0177 & 0.0133 & 0.2050 & 9.929 & 0.0460 \\
SITR~\cite{gupta2025sitr}
 & 0.0156 & 0.0118 & 0.0748 & 3.626 & 0.0214 \\
TVL~\cite{fu2024tvl}
 & 0.0210 & 0.0163 & 0.1057 & 5.084 & 0.0247 \\
Sparsh-DINOv2~\cite{higuera2024sparsh,oquab2023dinov2}
 & 0.0189 & 0.0139 & 0.1227 & 5.914 & 0.0268 \\
Sparsh-MAE~\cite{higuera2024sparsh,he2022mae}
 & 0.0172 & 0.0130 & 0.1255 & 6.030 & 0.0281 \\
\addlinespace[1.5pt]
TVL + MAE$^{\dag}$~\cite{fu2024tvl,he2022mae}
 & 0.0231 & 0.0160 & 0.0804 & 3.903 & 0.0216 \\
Sparsh-X+MAE$^{\dag}$~\cite{higuera2025sparshx,he2022mae}
 & 0.0199 & 0.0145 & 0.0503 & 2.411 & 0.0140 \\
\rowcolor{black!5}
\textbf{\method{}}$^{\dag}$
 & 0.0170 & 0.0128 & 0.0488 & 2.350 & 0.0137 \\
\midrule
\multicolumn{6}{@{}l}{\textit{(c) Both}} \\
VITaL (adapted)~\cite{george2025vital}
 & 0.0254 & 0.0182 & 0.0423 & 2.034 & 0.0112 \\
MViTac (adapted)~\cite{dave2024mvitac}
 & ---    & ---    & 0.0215 & 1.027 & 0.0072 \\
TVL + MAE$^{\dag}$~\cite{fu2024tvl,he2022mae}
 & 0.0106 & 0.0103 & 0.0143 & 0.720 & 0.0058 \\
Sparsh-X+MAE$^{\dag}$~\cite{higuera2025sparshx,he2022mae}
 & 0.0107 & 0.0101 & 0.0131 & 0.655 & 0.0055 \\
\rowcolor{black!5}
\textbf{\method{}}$^{\dag}$
 & \textbf{0.0066} & \textbf{0.0074} & \textbf{0.0129} & \textbf{0.652} & \textbf{0.0053} \\
\bottomrule
\end{tabularx}
\par\vspace{1pt}
\parbox{\columnwidth}{\footnotesize
Depth MSE is in mm$^2$. Rot.\ $L_1$ measures errors in
$(\cos\theta,\sin\theta)$; Trans.\ $L_1$ uses normalized coordinates.
Vis/Tac denote UV-off/on input; Both uses the pair.
$\dagger$: weights shared across input configurations within each selected checkpoint;
unmarked Vis/Tac rows are dedicated models. Geometry and pose use
separate task-selected checkpoints (Sec.~\ref{sec:setup}). Citations identify
source methods; all scores are our evaluations. TVL+MAE and Sparsh-X+MAE
are encoder substitutions, and Sparsh-X uses a \mbox{locally pretrained image encoder.}\unskip}
\caption{\textbf{Contact geometry and object pose on real validation data.} Bold: lowest error in the paired-input configuration; ties at displayed precision.\unskip}
\label{tab:main}
\end{table}
\subsection{Sensor Observations and Evaluation Protocol}
\label{sec:setup}
\noindent\textbf{Sensor observations and endurance}
Figure~\ref{fig:real_obj} illustrates common-view visual and tactile observations of four everyday objects. In the reported cyclic-loading test, a spherical indenter applied more than 45{,}000 normal indentations at 20\,N without observed cracking, delamination, or loss of the imaging-state change. This result characterizes the tested normal-loading protocol; sliding wear, contamination, and \mbox{optical drift were not quantified.}\unskip

\noindent\textbf{Data and geometric references}
A KUKA LBR Med R820 arm collects approximately 400 contacts for each of 12 objects, recording both image states and end-effector pose at every contact. A reference-object calibration relates object pose to the sensor frame. Pose and object geometry generate the depth reference maps; real normal maps are obtained by finite differences. These are computed geometric targets, not independent measurements of the deformed membrane. For the 12 corresponding geometries, Blender renders visual images, depth maps, and normal maps. Simulated tactile images are synthesized from the rendered depth maps, with random $\pm15\%$ depth scaling, by a depth-to-tactile generator (U-Net with a PatchGAN discriminator) trained only on real training pairs.\unskip

Both image streams are center-cropped to 0.816 of their original dimensions and resized to $224\times224$. Simulated visual images receive an additional $1.15\times$ center zoom to align their field of view with real images. Simulated poses are restricted to the per-object real rotation and translation ranges with a margin factor of 1.5. Simulation augmentation applies matched in-plane rotations within $\pm15^\circ$, together with channel, illumination, and noise perturbations. Real images are not augmented.

\noindent\textbf{Validation and checkpoint selection}
Every tenth sample within each real session is held out for validation and checkpoint selection. Objects and acquisition sessions are shared between training and validation, and object identity is supplied to the pose queries. All reported scores therefore measure within-session generalization on known objects.

Geometry and pose use separate checkpoints selected by depth MSE and mean angular error, respectively. Fusion models use both for selection; dedicated models use their native mode. Each selected fusion checkpoint is evaluated in all three modes without reselection, ensuring a common checkpoint for each within-task modality comparison.

\noindent\textbf{Geometry and pose metrics}
Depth and normal MSEs average over all pixels, including non-contact regions. Normal MSE also averages over its three components. Depth MSE is in $\mathrm{mm}^2$. Rotation $L_1$ sums the absolute cosine and sine errors per sample; angular error averages the inverse cosine of the clipped predicted--reference dot product, in degrees. Translation $L_1$ averages both coordinate errors, normalized by half the object's target size.

\noindent\textbf{Optimization}
The main fusion runs use real--simulation co-training at a real:sim ratio of 1:1.27. Optimization uses AdamW for 150 epochs with batch size 32, learning rate $2\times10^{-4}$, weight decay 0.05, cosine decay, and 1{,}000 warmup steps. We use mixed precision and gradient clipping at 1.0. The input probabilities are $(p_{\rm Vis},p_{\rm Tac},p_{\rm Both})=(0.05,0.25,0.70)$, with $p_{\rm inject}=0.75$.

\noindent\textbf{Compared estimators}
All scores in Table~\ref{tab:main} are local evaluations. TVL~\cite{fu2024tvl} and Sparsh-X~\cite{higuera2025sparshx} replace the tactile encoder while retaining MAE and our downstream architecture. The Sparsh-X image encoder is locally pretrained on the co-training images and then frozen. VITaL~\cite{george2025vital} and MViTac~\cite{dave2024mvitac} use adapted ResNet-18 implementations; MViTac predicts pose only. Dedicated single-modality models attach geometry and pose heads to one frozen encoder. Symmetric MBT~\cite{nagrani2021mbt} replaces the fusion trunk while retaining the same encoders and heads.

\subsection{Complementary Contributions of Vision and Touch}
\label{sec:baselines}
In Table~\ref{tab:main}, pairing vision and touch reduces \method{}'s depth MSE to $0.0066\,\mathrm{mm}^2$, corresponding to reductions of 93.7\% relative to Vis and 61.2\% relative to Tac. Angular error decreases by 72.3\%, from $2.350^\circ$ with Tac to $0.652^\circ$ with Both, and is also lower than the $0.694^\circ$ obtained with Vis. Vision therefore provides substantially stronger pose information than touch alone, while both modalities contribute to contact-depth estimation.

Among the paired-input models in Table~\ref{tab:main}, \method{} achieves the lowest error across all five metrics. Compared with Sparsh-X+MAE, it reduces depth MSE by 38.3\% and normal MSE by 26.7\%, while lowering translation $L_1$ from 0.0055 to 0.0053. The largest improvements occur in geometry, whereas angular errors remain similar at $0.652^\circ$ for \method{} and $0.655^\circ$ for Sparsh-X+MAE.

\begin{figure*}[t]
\centering
\includegraphics[width=0.92\textwidth]{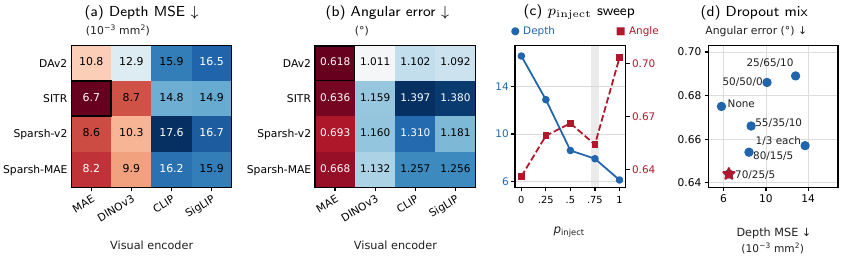}
\caption{\textbf{Encoder and training ablations with paired input.} (a,b) Depth MSE and angular error for four tactile encoders (rows) and four visual encoders (columns); outlines mark minima. (c) Visual-conditioning sweep with a fixed Both/Tac/Vis mixture of 55/35/10, digitized from the two-seed mean curve. Blue circles show depth MSE (left axis); red squares show angular error (right axis). (d) Modality mixtures at $p_{\rm inject}=0.75$; labels give Both/Tac/Vis percentages, and ``None'' denotes paired-only training. The star marks the selected mixture. Depth is in $10^{-3}\,\mathrm{mm}^2$ and angle in degrees. These sweeps are separate from the main-table run.}\label{fig:ablation_summary}
\end{figure*}

Dedicated models remain competitive with single input. SITR leads Tac depth and normal accuracy, Sparsh-X leads angular accuracy, and DAv2 leads translation accuracy. With Vis, T3 has the lowest depth, normal, and angular errors, while \method{} has the lowest translation error.

Modality dropout supports geometry and pose inference with either available input using shared parameters. Within \method{}, paired observations give the lowest error on all five metrics, combining this input flexibility with the benefit of complementary measurements.

\subsection{Representation and Training Design}
\label{sec:encoder_ablation}
\noindent\textbf{Encoder selection.} 
In Fig.~\ref{fig:ablation_summary}(a,b), MAE gives the lowest depth and angular errors for every tactile encoder. SITR+MAE achieves the lowest depth MSE of 0.0067, whereas DAv2+MAE achieves the lowest angular error of $0.618^\circ$. We select SITR+MAE to prioritize contact geometry. This sweep compares both pretrained representations and architectures.

\begin{table}[t]
\centering
\tableformat
\setlength{\tabcolsep}{2.4pt}
\begin{tabularx}{\columnwidth}{lLrrrrr}
\toprule
 & & \multicolumn{2}{c}{\textbf{Geometry}}
 & \multicolumn{3}{c}{\textbf{Pose}} \\
\cmidrule(lr){3-4} \cmidrule(lr){5-7}
Input & Model & \metrichead{Depth}{MSE} & \metrichead{Normal}{MSE}
 & \metrichead{Rot.}{$L_1$} & \metrichead{Angle}{$({}^\circ)$} & \metrichead{Trans.}{$L_1$} \\
\midrule
Vis & Sym. MBT~\cite{nagrani2021mbt}
 & 0.1059 & 0.0387 & 0.0168 & 0.831 & 0.0069 \\
\rowcolor{black!5}
 & \textbf{\method{}}
 & 0.1055 & 0.0388 & 0.0138 & 0.694 & 0.0058 \\
\midrule
Tac & Sym. MBT~\cite{nagrani2021mbt}
 & 0.0167 & 0.0123 & 0.0452 & 2.178 & 0.0120 \\
\rowcolor{black!5}
 & \textbf{\method{}}
 & 0.0170 & 0.0128 & 0.0488 & 2.350 & 0.0137 \\
\midrule
Both & Sym. MBT~\cite{nagrani2021mbt}
 & \textbf{0.0062} & \textbf{0.0071} & 0.0149 & 0.747 & 0.0054 \\
\rowcolor{black!5}
 & \textbf{\method{}}
 & 0.0066 & 0.0074 & \textbf{0.0129} & \textbf{0.652} & \textbf{0.0053} \\
\bottomrule
\end{tabularx}
\caption{\textbf{Fusion configuration comparison with SITR+MAE and identical heads.}
Bottleneck tokens: 4 (MBT), 16 (ours). Checkpoints as in Table~\ref{tab:main}; bold: lower error in the paired-input comparison.\unskip}
\label{tab:arch}
\end{table}

\vspace{2pt}
\noindent\textbf{Fusion configuration.}
Using the same encoders and heads, Table~\ref{tab:arch} shows that asymmetric fusion reduces paired-input angular error from $0.747^\circ$ to $0.652^\circ$ and translation $L_1$ from 0.0054 to 0.0053. Symmetric MBT achieves lower depth and normal MSEs and the lowest errors across all five Tac metrics. These results characterize a geometry--pose trade-off between the complete fusion configurations: asymmetric fusion uses 16 bottleneck tokens, whereas symmetric MBT uses four.

\vspace{2pt}
\noindent\textbf{Computational cost}
\method{} has 92.5\,M trainable parameters, whereas symmetric MBT has 83.0\,M trainable parameters out of 482.9\,M total. The fusion trunks account for the 9.5\,M difference in trainable parameters.

\begin{figure}[h]
\centering
\includegraphics[width=\columnwidth]{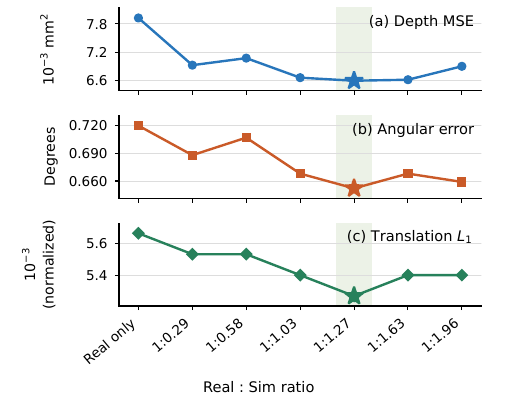}
\caption{\textbf{Real--simulation data mixing with paired input.} Completed configurations are compared by (a) depth MSE, (b) mean angular error, and (c) normalized translation $L_1$. Stars mark the main-experiment ratio of 1:1.27. Depth uses the depth-selected checkpoint; both pose metrics use the angle-selected checkpoint.}
\label{fig:data_mixing}
\end{figure}

\vspace{2pt}
\noindent\textbf{Data mixing and modality dropout}
Figure~\ref{fig:data_mixing} compares the completed real:sim ratio configurations. At 1:1.27, the main-experiment configuration gives depth MSE 0.0066, angular error $0.652^\circ$, and translation $L_1$ 0.0053, compared with 0.0079, $0.720^\circ$, and 0.0056 for real-only training. The 1:1.63 configuration also reaches depth MSE 0.0066 at the reported precision; 1:1.27 gives the lowest angular and translation errors among the plotted configurations. Real-only and mixed-data runs use 100 and 150 epochs, respectively, with the number of updates also depending on dataset size.

Figure~\ref{fig:ablation_summary}(c) varies visual conditioning while fixing the modality probabilities at 0.55 for Both, 0.35 for Tac, and 0.10 for Vis. Depth MSE decreases with more frequent conditioning, whereas angular error varies nonmonotonically. With $p_{\rm inject}=0.75$, panel~(d) compares different modality mixtures. The selected mixture assigns probabilities of 0.70, 0.25, and 0.05 to Both, Tac, and Vis, respectively, yielding a depth MSE of 0.0065 and an angular error of $0.644^\circ$. Paired-only training lowers depth MSE to 0.0058 but increases angular error to $0.675^\circ$. These sweeps evaluate paired-input accuracy, while Table~\ref{tab:main} separately assesses robustness to missing modalities.

\begin{figure*}[h]
\centering
\includegraphics[width=\textwidth]{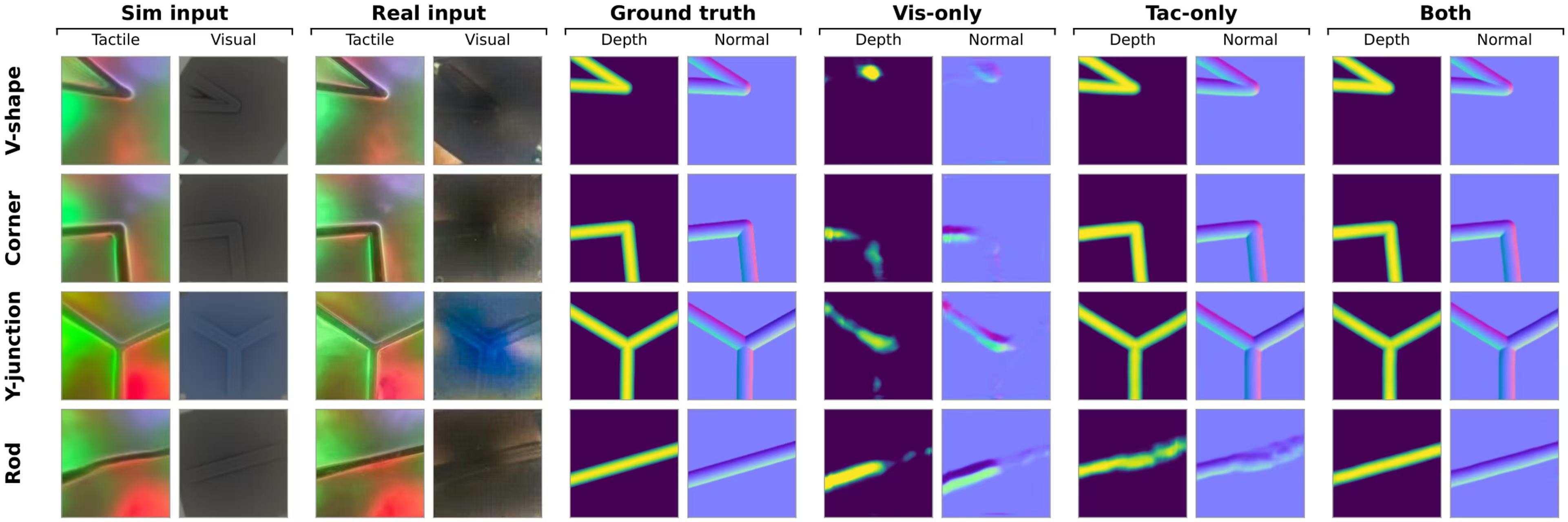}
\caption{\textbf{Depth and normal predictions on real validation samples.} Sim inputs are shown for comparison: tactile images are generated from each sample's reference depth, and visual images are rendered at the corresponding pose. Predictions use the real inputs and one depth-selected checkpoint across all three modes. Depth uses a shared color range within each row, and normals are RGB-encoded.}
\label{fig:qualitative}
\end{figure*}

\noindent\textbf{Qualitative geometry}
Figure~\ref{fig:qualitative} shows errors that full-image averages do not localize. Vis-only predictions miss much of the V-shape and corner contact regions; tactile input recovers their boundaries. Paired input also reduces the irregular rod depth profile visible with Tac alone.

\section{DISCUSSION AND FUTURE WORK}
\label{sec:conclusion_future_work}

\subsection{Discussion}

We presented \sensor{} and \method{} as a common-view visual--tactile sensing and estimation framework for object pose and contact geometry.

Our results suggest that the value of multimodal sensing lies not only in the distinct strengths of individual modalities, but also in their ability to resolve each other's ambiguities. Proximity vision provides a global prior for object pose and contact interpretation, whereas tactile sensing supplies a direct local measurement of contact geometry. The further improvement obtained with paired observations indicates cross-modal synergy rather than a simple division of sensing roles, motivating task-specific asymmetric fusion and state-dependent mode selection for dexterous manipulation.

Our real--simulation co-training study further reveals a trade-off between synthetic-data coverage and domain fidelity. Adding simulated observations expands the range of object poses and contact geometries seen during training and improves real-world validation performance over real-only training. However, residual sim--real discrepancies---including differences in illumination, blur, and background appearance in proximity mode that are difficult to eliminate---can introduce simulation-specific cues. As simulated samples account for a larger fraction of training, these discrepancies may increasingly bias the learned representation away from the real-data distribution, resulting in negative transfer and an intermediate optimal mixing ratio rather than monotonic gains from additional simulation. In our setting, the selected real-to-simulation ratio of 1:1.27 provides the best overall trade-off for pose estimation while retaining competitive geometry accuracy. This optimum is specific to the current simulator, dataset, and training configuration; more broadly, the result suggests that effective sim--real co-training must balance synthetic coverage against domain fidelity.

\subsection{Future Work}
\label{sec:limitations}

\noindent\textbf{Miniaturization and thermal management}
The current prototype retains the GelSlim~4.0 body; future work will co-design the camera--lens module, illumination, electronics, and contact stack for a smaller, power-efficient fingertip while preserving field of view and contact sensitivity. We will quantify package size, energy per paired observation, and contact-surface temperature under matched acquisition conditions.\unskip

\noindent\textbf{Perception evaluation}
Future work will evaluate cross-session generalization and validate local contact geometry using contact-region metrics and direct membrane-shape measurements.\unskip

\end{document}